\documentclass[letterpaper, 10 pt, conference]{ieeeconf}  

\IEEEoverridecommandlockouts                              

\usepackage{graphicx}
\usepackage{booktabs}
\usepackage{mathtools}
\usepackage{amsmath}
\usepackage{multicol}
\usepackage{multirow}
\usepackage{soul}
\usepackage{float}
\usepackage{placeins}
\usepackage[dvipsnames]{xcolor}
\let\labelindent\relax
\usepackage{enumitem}
\usepackage{tabularx}
\usepackage{cite}
\usepackage{hyperref}
\usepackage{amssymb}
\usepackage{subcaption}
\hypersetup{
    colorlinks=true,
    linkcolor=blue,
    filecolor=magenta,      
    urlcolor=cyan,
    pdftitle={Overleaf Example},
    pdfpagemode=FullScreen,
    }
\hypersetup{hidelinks}
\usepackage{dsfont}
\usepackage{cleveref}
\title{\LARGE \bf
Towards Active Excitation-Based \\ Dynamic Inertia Identification in Satellites}

\author{Matteo El Hariry$^{1}$, Vittorio Franzese$^{1}$, Miguel Olivares-Mendez$^{1}$
\thanks{$^{1}$SnT, University of Luxembourg, 29 Av. John F. Kennedy, 1855 Kirchberg Luxembourg.
        E-mails: matteo.elhariry@uni.lu ; vittorio.franzese@uni.lu ; miguel.olivaresmendez@uni.lu}%
}

\begin{document}

\maketitle
\thispagestyle{empty}
\pagestyle{empty}

\begin{abstract}
This paper presents a comprehensive analysis of how excitation design influences the identification of the inertia properties of rigid nano- and micro-satellites. We simulate nonlinear attitude dynamics with reaction‑wheel coupling, actuator limits, and external disturbances, and excite the system using eight torque profiles of varying spectral richness.
Two estimators are compared, a batch Least Squares method and an Extended Kalman Filter, across three satellite configurations and time‑varying inertia scenarios. Results show that excitation frequency content and estimator assumptions jointly determine estimation accuracy and robustness, offering practical guidance for in‑orbit adaptive inertia identification by outlining the conditions under which each method performs best. The code is provided as open-source\footnote{\label{fn:code}\url{https://github.com/elharirymatteo/satellite-inertia-id}}.
\end{abstract}

\section{Introduction}

The ability to accurately characterize the inertia properties of a spacecraft while in mission flight is key for attitude and trajectory control. Inertia tensors govern the rotational response of a satellite to control inputs, and mismatches between assumed and actual values can lead to degraded performance or instability in closed-loop systems. In practice, these parameters often deviate from pre‑launch models due to configuration changes or fuel depletion. This motivates on‑board, data‑driven identification using only the actuation and sensing available during flight.

In this work, we investigate how the temporal structure of applied control torques affects the identifiability and accuracy of spacecraft inertia estimates. Using orthogonally aligned reaction wheels (RWs) to inject known torque inputs and observing the resulting body-frame angular velocities, we systematically analyze the role of excitation design in shaping the quality of recovered inertia parameters. Unlike studies that focus on proposing or optimizing estimation algorithms, our contribution lies in conducting a broad comparative analysis across excitation patterns, estimator types, and satellite configurations.

We begin by detailing the simulation setup and excitation design used to evaluate identifiability. We then compare the performance of LS and EKF across different satellites and dynamic inertia conditions. This allows us to characterize the conditions under which each method proves most effective and to highlight key trade-offs for in-orbit application.

We consider two estimators: a batch least-squares (LS) regression method assuming static inertia, and an extended Kalman filter (EKF) capable of tracking time-varying parameters. Both are tested across three representative satellite configurations, incorporating reaction wheel dynamics, actuation constraints, and external disturbances. For dynamic-inertia scenarios, motivated by events such as fuel consumption or deployment mechanisms, we extend the simulation environment to support step changes, slow drifts, and oscillatory variations in inertia, enabling controlled studies of estimator behavior under nonstationary conditions.

The torque profiles tested, summarized in Table~\ref{tab:torque_profiles}, range from smooth sinusoids to discontinuous binary sequences, covering a broad spectrum of excitation richness. We benchmark them in terms of estimation error and regression conditioning, analyzing not only differences between LS and EKF, but also how these interact with excitation design, satellite properties, and dynamic inertia changes. The resulting analysis provides insights into how excitation design and estimator choice jointly influence identification outcomes, offering guidance for the design of future in-orbit inertia estimation strategies.
To support reproducibility and further experimentation, all code and simulation scenarios are released as open source at \footref{fn:code}.


\section{Related Work}

Estimating spacecraft inertia has long been addressed through sensor‑based system identification, yet most prior works emphasize estimator formulation rather than input‑excitation design. Classical approaches rely on gyro-based measurements combined with commanded torques. Early works introduced online recursive least-squares algorithms using gyroscopes to identify the center of mass and inertia tensor of a spacecraft~\cite{Wilson2002gyro}. This concept has also been augmented with the conservation of angular momentum to derive mass properties from attitude and rate data~\cite{Meng2020mass}. Similar methods based on least-squares formulations with physical constraints have been explored by various authors~\cite{Bergmann1987mass, bordany2000orbit, cheriet2021inertia}.

Recursive filtering techniques such as the Extended Kalman Filter and Unscented Kalman Filter have been used to estimate the spacecraft angular rates from gyroscope data and applied torques, considering inertia uncertainty ~\cite{bellar2019satellite, norman2011orbit, cao2016unscented}. Iterative filtering and batch-hybrid methods have also been proposed, where an EKF estimates states while a separate batch algorithm refines the inertia matrix until convergence~\cite{kim2010combined}. More recently, machine learning approaches have also emerged as an alternative solution to the inertial matrix estimation problem as deep neural network framework~\cite{wu2024inertia} has been used to improve accuracy in the presence of complex measurement noise. Several works have also addressed challenges of estimating platform with a non-diagonal inertia matrix and external disturbances~\cite{soken2014estimation}. 

Despite extensive literature on LS and filtering approaches, little research has systematically compared how excitation profiles influence estimator performance across satellites and varying inertia conditions. 
This gap motivates the comparative framework developed in this study.

Recent works have explored the use of deep reinforcement learning (DRL) for robust spacecraft attitude control in highly nonlinear and underactuated settings, such as reaction wheel failures~\cite{hariry2025deep, elkins2020autonomous, xu2019model}. While our focus lies on adaptive identification of time-varying inertia properties, such estimation strategies can be integrated with DRL controllers to enable resilient and quasi-optimal attitude regulation under uncertain and evolving spacecraft conditions.


\section{Preliminaries}
\subsection{Attitude Dynamics}
The rotational dynamics of a satellite follows the conservation of the angular momentum. In presence of a triad of orthogonal reaction wheels, the total angular momentum $\mathbf{h}$ of a satellite reads 
\begin{equation}
\mathbf{h} = \mathbf{I}_s \boldsymbol{\omega} + \mathbf{h}_{\text{rw}}
\end{equation}
where $\mathbf{I}_s$ is the inertia tensor of the spacecraft, $\boldsymbol{\omega}$ is the spacecraft angular rate vector, and
\begin{equation}
\mathbf{h}_{\text{rw}} = \sum_{i=1}^{3} I_{\text{rw}} \omega_{\text{rw},i} \hat{\mathbf{e}}_i
\end{equation}
is the angular momentum contribution of the reaction wheels, where $I_{\text{rw}}$ is the inertia of the reaction wheels along their spin axis, assumed to be identical for the three wheels, and $\omega_{\text{rw},i}$ is the rotational speed of the $i$-th wheel with spin axis $\hat{\mathbf{e}}_i$. In presence of constant inertia, the conservation of the angular momentum leads to the Euler’s rigid body equations
\begin{equation}
\mathbf{I}_s \dot{\boldsymbol{\omega}} = -\boldsymbol{\omega} \times \left( \mathbf{I}_s \boldsymbol{\omega} + \mathbf{h}_{\text{rw}} \right) - \dot{\mathbf{h}}_{\text{rw}} + \boldsymbol{\tau}_{\text{ext}}
\end{equation}
which accounts for the internal momentum exchange from reaction wheels and the presence of external torques $\boldsymbol{\tau}_{\text{ext}}$ such as gravity gradient or solar radiation pressure. The term
$\dot{\mathbf{h}}_{\text{rw}}$ accounts for the acceleration of the reaction wheels due to applied control torques. This system can be numerically integrated using a fixed-step explicit Runge–Kutta (as RK45 or RK78) method from the \texttt{SciPy} library~\cite{scipy2020}, with torque input provided over time.

\begin{figure}[ht]
    \centering
    \includegraphics[width=0.48\textwidth]{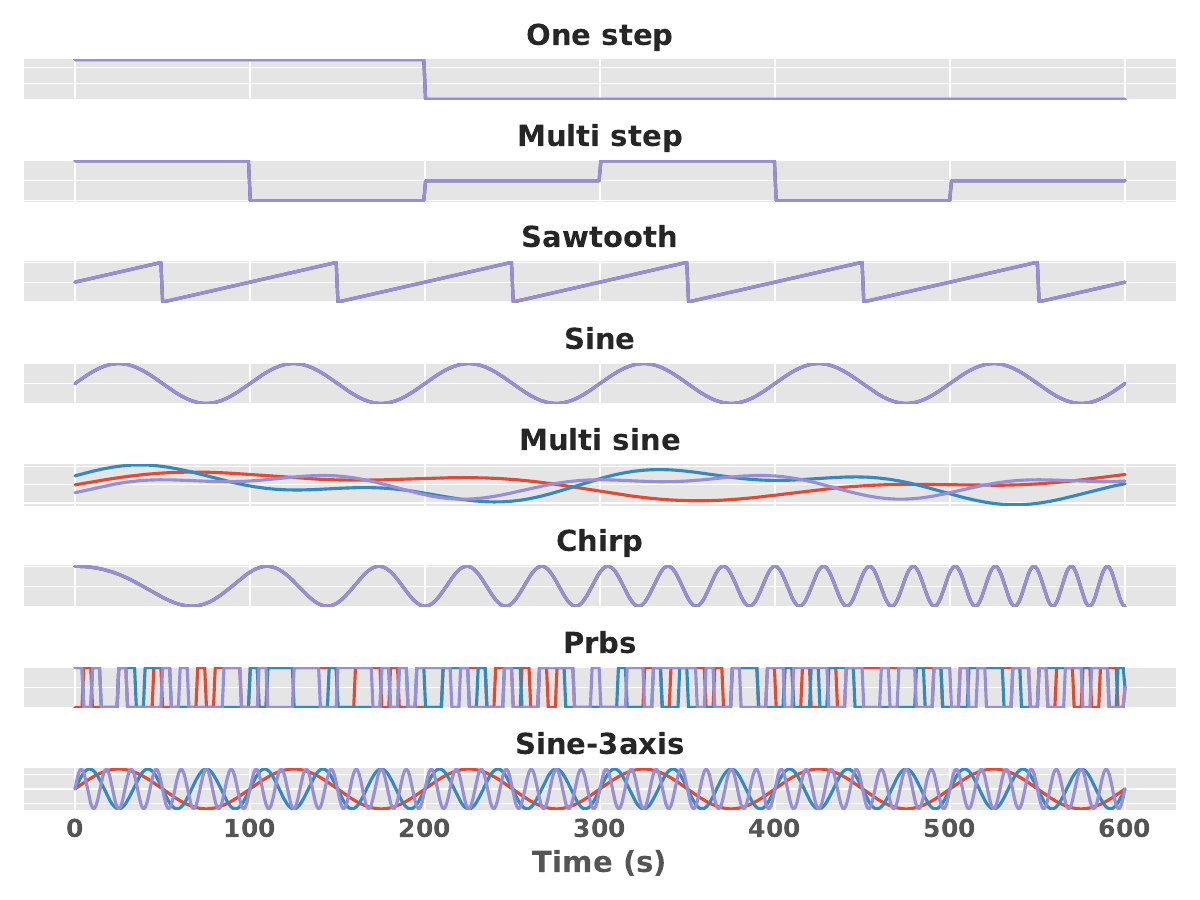}
    \caption{
    \small
        Overview of torque excitation profiles used for system identification. The figure shows eight profiles plotted as simultaneous signals:
        \textit{One step}, \textit{Multi step}, \textit{Sawtooth}, \textit{Sine}, \textit{Multi-sine}, \textit{Chirp}. \textit{PRBS}, and \textit{Sine 3-axis}. 
        These profiles have been designed to maximize parameter observability in different degrees, varying their frequency content and excitation characteristics.
    }
    \label{fig:torque_profiles}
\end{figure}

\begin{table}[t]
\centering
\caption{Torque profiles used to excite satellite dynamics. Each profile is normalized and applied per axis.}
\label{tab:torque_profiles}
\begin{tabular}{llll}
\toprule
\textbf{No.} & \textbf{Name} & \textbf{Type} & \textbf{Description} \\ \midrule
1 & \textbf{one step}      & Discrete     & Single impulse per axis \\
2 &\textbf{multi step}           & Discrete     & Alternating step changes \\
3 &\textbf{sawtooth}           & Discrete     & Alternating linear ramps \\
4 &\textbf{sine}           & Periodic     & Single-frequency sinusoid \\
5 &\textbf{multi sine}    & Periodic     & Sum of multiple harmonics \\
6 &\textbf{chirp}          & Periodic     & Frequency-sweeping sine \\
7 &\textbf{prbs}           & Binary       & Pseudo-random binary input \\
8 &\textbf{sine 3-axis}         & Periodic   & Multi-frequency sinusoids \\

\bottomrule
\end{tabular}
\end{table}

\subsection{Torque Profile Generation}
Through the use of reaction wheels we generate a variety of torque profiles $\mathbf{\dot{h}_{rw}}$ to explore how different control inputs affect the identification of the spacecraft inertia parameters. These include smooth, band-limited, and discontinuous signals, selected to span a spectrum of temporal structures and excitation bandwidths. The full set of input types is summarized in Table~\ref{tab:torque_profiles} and shown in Figure~\ref{fig:torque_profiles}. Profiles are normalized to respect actuator saturation limits and applied through reaction wheels as body torques. Each profile yields to a different level of information according to the amount of excitation it can provide.

\begin{table*}[ht!]
\caption{Physical and reaction wheel parameters of the three satellite models.}
\label{tab:sat_params}
\centering
\small
\setlength{\tabcolsep}{6pt}
\begin{tabular}{@{}lccccccc@{}}
\toprule
\textbf{Satellite} & \textbf{Mass} & \textbf{Dimensions} & \textbf{Inertia Diagonal} & \textbf{RW Max Torque} & \textbf{RW Diameter} & \textbf{RW Max Speed} & \textbf{RW Inertia} \\
 & [kg] & [m] & [kg$\cdot$m$^2$] & [Nm] & [m] & [rad/s] & [kg$\cdot$m$^2$] \\
\midrule
CubeSat & 24.0 & 0.2$\times$0.2$\times$0.3 &
[0.26, 0.26, 0.16] & 0.01 & 0.06 & 460 & $1.0\times10^{-4}$ \\
Microsat & 95.0 & 0.5$\times$0.6$\times$0.8 & [6.53, 5.96, 4.53] & 0.1 & 0.12 & 900 & $2.0\times10^{-3}$ \\
Small Sat & 118.0 & 0.7$\times$0.8$\times$1.0 & [10.6, 14.2, 15.3] & 0.1 & 0.14 & 1500 & $3.0\times10^{-3}$ \\
\bottomrule
\end{tabular}
\vspace{-.4cm}
\end{table*}

\subsection{Least Squares Estimation}
In the LS method, we estimate the inertia tensor of the satellite from time series of applied torques and measured angular velocities. We can note that, under exact parameters and state variables knowledge, the following equations holds:
\begin{equation}
\boldsymbol{0} = \mathbf{I}_s \dot{\boldsymbol{\omega}} + \boldsymbol{\omega} \times \left( \mathbf{I}_s \boldsymbol{\omega} + \mathbf{h}_{\text{rw}} \right) + \dot{\mathbf{h}}_{\text{rw}} - \boldsymbol{\tau}_{\text{ext}}
\end{equation}
Under uncertain parameters and state variables, however, we can define the residual:
\begin{equation}
\boldsymbol{\varepsilon} = \mathbf{I}_s \dot{\boldsymbol{\omega}} + \boldsymbol{\omega} \times \left( \mathbf{I}_s \boldsymbol{\omega} + \mathbf{h}_{\text{rw}} \right) + \dot{\mathbf{h}}_{\text{rw}} - \boldsymbol{\tau}_{\text{ext}}
\end{equation}
where $\dot{\boldsymbol{\omega}}$ is estimated using finite differences over the spacecraft angular rate, which is measured by gyroscopes, and inertia properties are uncertain. In this way, we can formulate a least squares optimization minimizing the function
\begin{equation}
J = \sum_t \boldsymbol{\varepsilon}^\top{\boldsymbol{\varepsilon}}
\end{equation} 
over the simulation time window. The optimization is solved using bounded trust-region methods to ensure physical feasibility of the inertia values.

\subsection{Extended Kalman Filter Estimation}

To enable online tracking of time-varying inertia, we formulate an extended Kalman filter that jointly estimates the spacecraft angular velocity $\boldsymbol{\omega} \in \mathbb{R}^3$, the principal moments of inertia $\mathbf{I}_s \in \mathbb{R}^3$, and the reaction wheel speeds $\boldsymbol{\omega}_{\text{rw}} \in \mathbb{R}^3$. The state vector is defined as $\mathbf{x} = [\boldsymbol{\omega}^\top, \mathbf{I}_s^\top, \boldsymbol{\omega}_{\text{rw}}^\top]^\top \in \mathbb{R}^9$, and the control input is the vector of reaction wheel accelerations $\dot{\boldsymbol{\omega}}_{\text{rw}}$. The continuous-time dynamics follow a simplified form of Euler’s equations with explicitly modeled control and reaction wheel kinematics. Assuming diagonal inertia and full observability of $\boldsymbol{\omega}$ and $\boldsymbol{\omega}_{\text{rw}}$, the nonlinear dynamics are discretized using a first-order Euler scheme to derive a transition function $\mathbf{x}_{k+1} = f(\mathbf{x}_k, \mathbf{u}_k) + \mathbf{w}_k$, where $\mathbf{w}_k$ is a zero-mean Gaussian process noise. The inertia terms $\mathbf{I}_s$ are treated as a random walk process, reflecting slow but non-negligible variations in mass distribution over time. The EKF prediction and update steps are standard, with linearization performed via Jacobians computed around the current estimate. The measurement model assumes noisy access to $\boldsymbol{\omega}$ and $\boldsymbol{\omega}_{\text{rw}}$, yielding an observation vector $\mathbf{z}_k = h(\mathbf{x}_k) + \mathbf{v}_k$, where $\mathbf{v}_k$ is measurement noise with covariance $\mathbf{R}$. The state covariance matrix $\mathbf{P}_k$ is propagated using the Joseph form to ensure numerical stability and symmetry preservation. Therefore, $\mathbf{P}_{k+1}^+$ = ($\mathbf{I}$-$\mathbf{K}_k \mathbf{H}_k$)$\mathbf{P}_{k+1}^-$($\mathbf{I} $-$\mathbf{K}_k \mathbf{H}_k$)$^\top$ + $\mathbf{K}_k \mathbf{R}_k \mathbf{K}_k^{\top}$, where $\mathbf{P}_{k+1}^+$ is the updated covariance, $\mathbf{P}_{k+1}^+$ is the predicted covariance, $\mathbf{K}_k$ is the Kalman gain, $\mathbf{H}_k$ is the measurement function Jacobian, $\mathbf{R}_k$ is the measurement covariance matrix, and $\mathbf{I}$ is the identity matrix. \\
While both LS and EKF estimators operate without enforcing physical constraints such as the triangle inequalities on the principal moments of inertia, all simulations start from valid initial conditions. In our experiments, no violations were observed. Nevertheless, enforcing such constraints via projection or constrained filtering could improve robustness in noisy or long-horizon scenarios.

\begin{figure*}[!t]
\small
    \centering
    \includegraphics[width=0.39\textwidth, height=.22\textheight]{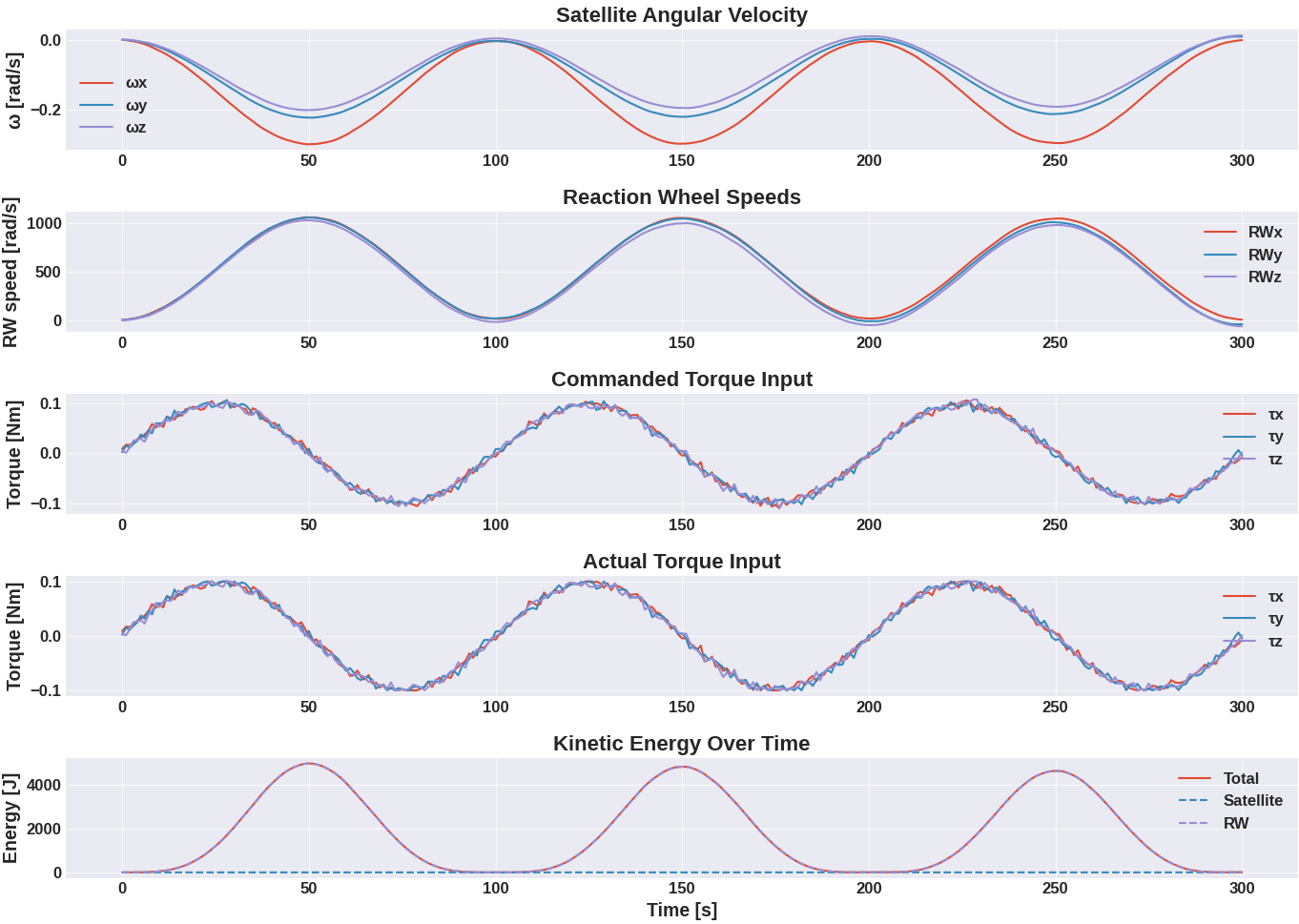}
    \includegraphics[width=0.60\textwidth, height=.22\textheight]{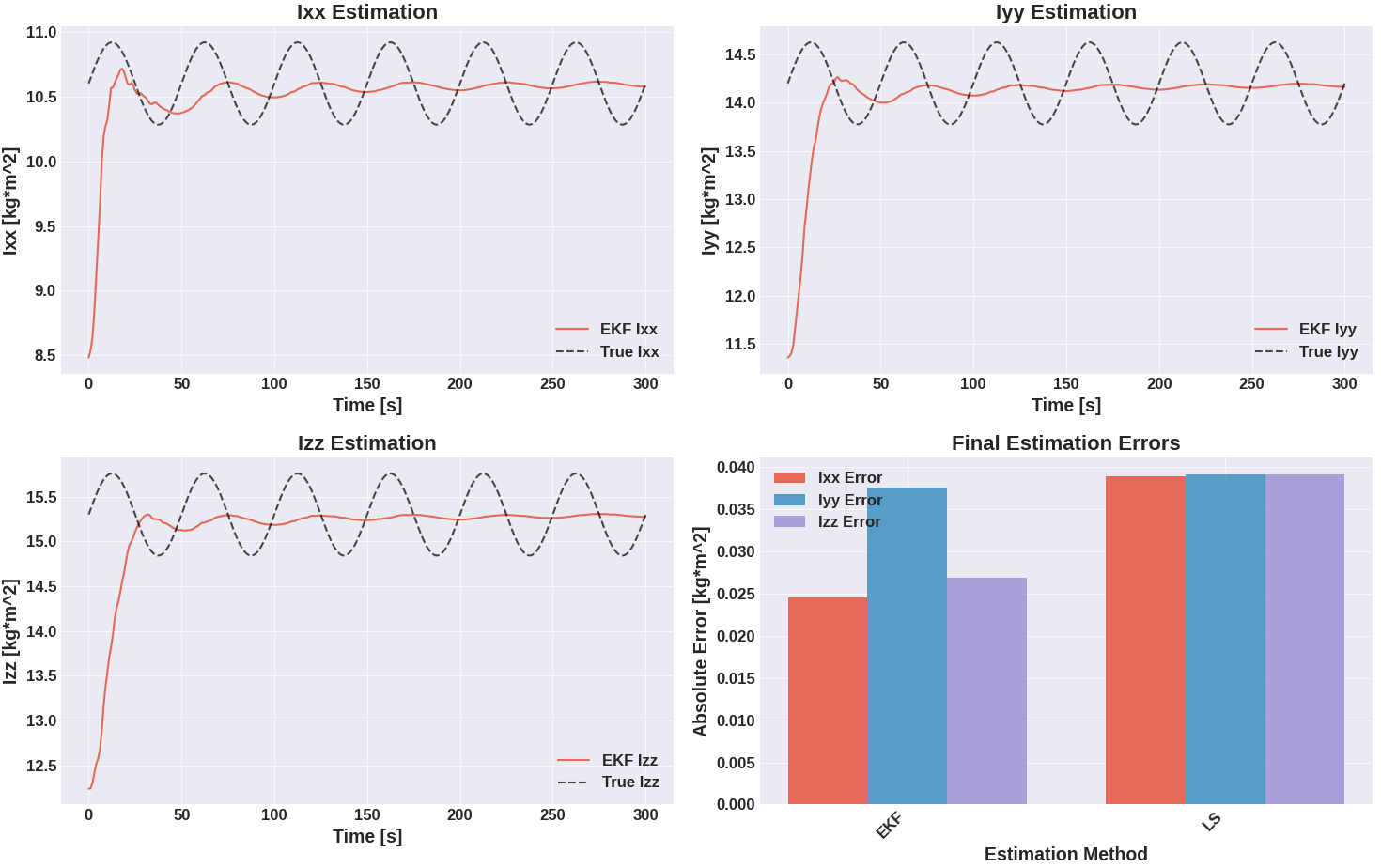} \\
    \includegraphics[width=0.8\textwidth, height=.14\textheight]{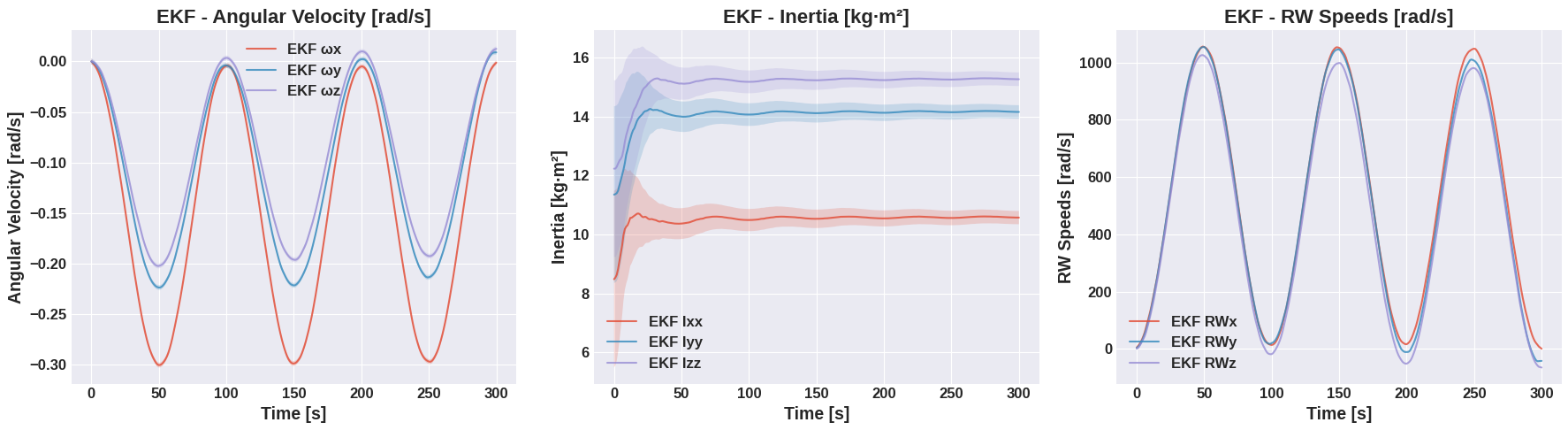}
    \caption{
    \textbf{Simulation and estimation pipeline overview.}
    (\textit{Top left}) Example of satellite dynamics and actuation under a sinusoidal torque profile: body-frame angular velocity, reaction wheel speeds, commanded vs. actual applied torques, and total/split kinetic energy evolution.
    (\textit{Top right}) Axis-wise EKF inertia tracking against ground truth for a time-varying inertia scenario, along with final absolute estimation errors compared to the LS method.
    (\textit{Bottom}) EKF tracking performance for angular velocity, estimated inertia components, and reaction wheel speeds over time.
        Together, these plots illustrate (i) the excitation characteristics and physical variables simulated, (ii) the EKF’s capability to track both kinematic states and inertia parameters, and (iii) the relative estimation accuracy across methods.
    }
    \label{fig:sim_pipeline_overview}
    \vspace{-.4cm}
\end{figure*}
\section{Methods}

\subsection{Simulation Framework}

We use a modular Python simulation that models a 3D rigid-body spacecraft equipped with a triad of orthogonally aligned reaction wheels. To explore estimator behavior under varying physical conditions, we consider three spacecrafts of increasing scale: a CubeSat, a Microsatellite, and a Small Satellite. The main physical and actuator characteristics of these spacecraft are summarized in Table~\ref{tab:sat_params}.

The simulator implements nonlinear rigid-body dynamics with realistic actuator, sensor, and disturbances. Reaction wheels are modeled with torque and speed saturation, such that the applied torque deviate from the commanded input depending on actuator limits. Torque bounds range from $0.01$ to $0.01\,\mathrm{Nm}$, with maximum speeds up to $1500\,\mathrm{rad/s}$ depending on the satellite configuration. Angular velocity measurements are disturbed with additive Gaussian noise, simulating gyroscope imperfections. External torques from gravity gradient and solar radiation pressure are included using simplified models based on a fixed orbital rate ($0.001\,\mathrm{rad/s}$), small force coefficients ($10^{-6}\,\mathrm{N}$), and a $1\,\mathrm{cm}$ center-of-pressure offset. While simplified, these effects introduce non-idealities that help evaluate estimator robustness.

To examine behavior under nonstationary conditions, we also simulate time-varying inertia using three physically plausible models: (i)~a \textit{step change} in inertia magnitude halfway through the episode (e.g., partial deployment), (ii)~a \textit{linear drift} representing slow structural adaptation or fuel depletion, and (iii)~a \textit{periodic variation} to emulate mass shifts due to fluid slosh or internal translation. The true inertia tensor $\mathbf{I}_s(t)$ is modified within the simulation, while the estimators are not informed of its variability. Let $\mathbf{I}_0$ be the nominal diagonal inertia and $t_{\max}$ the episode duration. We consider three models for the time-varying $\mathbf{I}_s(t)$:
\begin{align}
\text{(Step change)} \quad 
& \mathbf{I}_s(t) =
\begin{cases}
\mathbf{I}_0, & t < \tfrac{t_{\max}}{2}, \\[0.3em]
1.1\,\mathbf{I}_0, & t \ge \tfrac{t_{\max}}{2},
\end{cases} \\[0.8em]
\text{(Linear drift)} \quad 
& \mathbf{I}_s(t) = \mathbf{I}_0 \left( 1 + 0.05\,\tfrac{t}{t_{\max}} \right), \\[0.8em]
\text{(Periodic variation)} \quad 
& \mathbf{I}_s(t) = \mathbf{I}_0 \left[ 1 + 0.03\,\sin\!\left( 2\pi f\, t \right) \right],
\end{align}
with $f = 0.02~\mathrm{Hz}$ in the periodic case. 
The scaling factors---$5\%$ linear drift, $3\%$ sinusoidal amplitude, and $10\%$ step increase---are 
chosen to represent realistic in-orbit variations without introducing unphysical dynamics.

To illustrate the complete simulation and estimation process, Figure~\ref{fig:sim_pipeline_overview} presents an example run using a sinusoidal excitation profile on the Microsatellite configuration. The states  over time of the spacecraft and actuator are shown along with commanded and applied torques. The middle row compares estimated and true inertia values in a time-varying scenario, together with the final estimation errors for both EKF and LS, while the bottom row depicts the EKF’s ability to track angular velocity, reaction wheel speeds, and inertia components. This example provides a visual overview of the physical processes simulated and the types of outputs analyzed in the subsequent results section.

\subsection{Estimation and Evaluation Pipeline}

A torque excitation profile is selected from a predefined set of eight normalized signals and scaled to respect actuator constraints. The spacecraft is simulated for a fixed horizon (300\,s), including actuator dynamics, external torques, and, when applicable, time-varying inertia. Angular velocities and reaction wheel speeds are measured with additive Gaussian noise.
Two estimators are applied: (1) A \textbf{batch least-squares} method that assumes constant inertia and reconstructs angular accelerations via central differences. (2) an \textbf{extended Kalman filter} that models inertia as a random-walk state and jointly estimates angular velocity, reaction wheel speed, and inertia from noisy measurements and control inputs.  
Both methods are evaluated under identical simulation conditions to allow for direct comparison. The goal is not to declare a single best estimator, but to examine their respective strengths, limitations, and sensitivities to excitation profiles, spacecraft properties, and inertia variability. All experiments are repeated with 10 seeds per setting, and performance metrics are aggregated to report both mean and standard deviation of error.
In summary, LS accuracy depends primarily on the global conditioning of the regression, whereas EKF accuracy depends on the temporal richness of the excitation that drives informative state updates.

\subsection{Evaluation Metrics}
\begin{figure*}[!ht]
    \centering    \includegraphics[width=1.\textwidth, height=0.2\textheight]{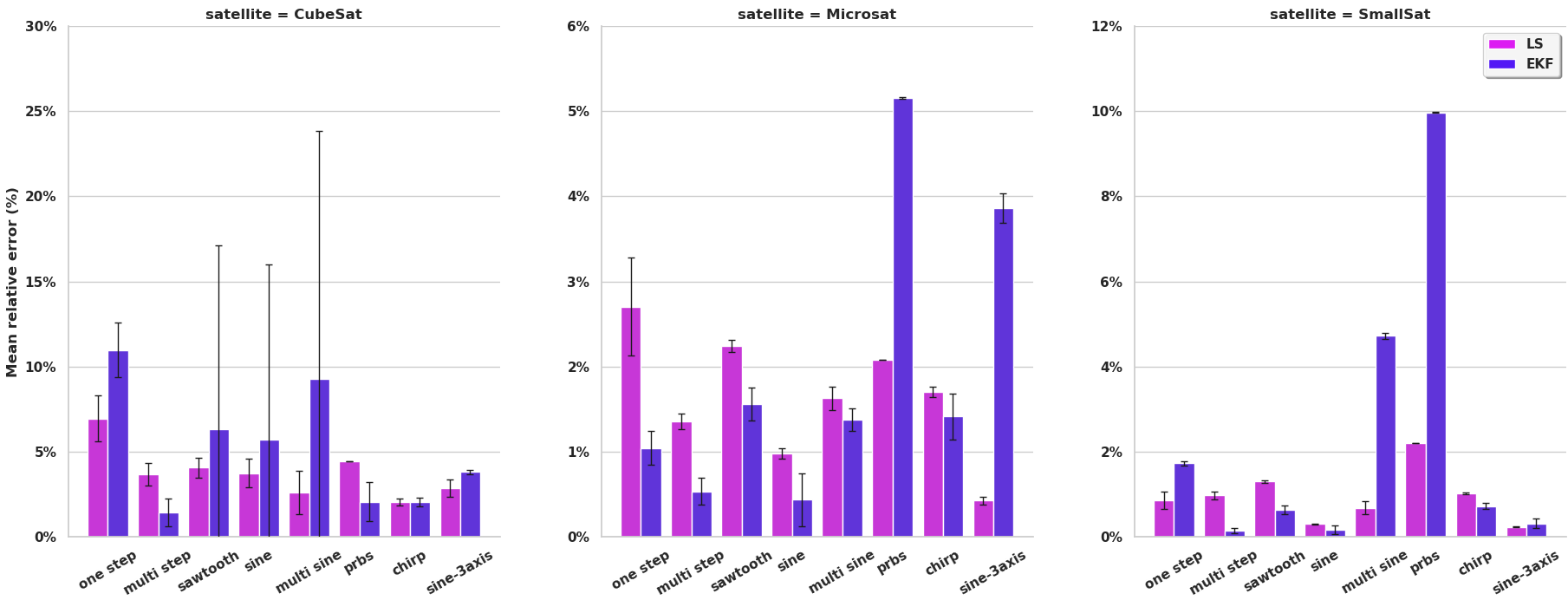}
\caption{Mean and std deviation of inertia estimation for static inertia scenario, grouped by satellite configuration and excitation profile. Error bars represent variability across seeds, highlighting conditions where each estimator performs more reliably.}
    \label{fig:static_barplots}
    \vspace{-0.4cm}
\end{figure*}
Estimator performance is quantified using:

\begin{itemize}
    \item \textbf{Normalized Error} (LS/EKF): The global estimation error is computed as the Euclidean norm of the difference between estimated and ground-truth inertia, normalized by the magnitude of the true inertia:
    \begin{equation}
        \text{Error} = \frac{\lVert \hat{\mathbf{I}}_s - \mathbf{I}_s^{\text{true}} \rVert_2}{\lVert \mathbf{I}_s^{\text{true}} \rVert_2}
    \end{equation}
    
    \item \textbf{Sliding-Window Normalized Error} (EKF, dynamic inertia): 
    For scenarios with time-varying inertia, the error is averaged over the last $K$ timesteps to capture steady-state estimation accuracy:
    \begin{equation}
        \text{Error}_{\text{dyn}} = \frac{1}{K} \sum_{t = t_{\max} - K + 1}^{t_{\max}} \frac{\lVert \hat{\mathbf{I}}_s(t) - \mathbf{I}_s^{\text{true}}(t) \rVert_2}{\lVert \mathbf{I}_s^{\text{true}}(t) \rVert_2}
    \end{equation}
    This approach mitigates transient effects and focuses on the final performance regime.
\end{itemize}

For static-inertia experiments, metrics are computed at $t = t_{\max}$. For dynamic-inertia scenarios, the reference value is computed over the final portion of the trajectory, using the last $K$ timesteps (20\% of $t_{\max}$). This sliding-window formulation provides a robust measure of steady-state performance by averaging the relative norm difference between the estimated and instantaneous ground-truth inertias.

\section{Experiments and Results}

We present a series of numerical-simulation experiments to assess the accuracy, robustness, and adaptability of inertia estimation methods under varying excitation profiles, satellite configurations, and inertia dynamics. Results are organized by estimation approach, input profile, and inertia model.
\begin{table}[!t]
\centering
\caption{Mean relative estimation error (\%) for LS/EKF on static inertia. Best method per profile/satellite is in \textbf{bold}.}
\label{tab:static_inertia_grouped}
\small
\begin{tabular}{lcccccc}
\toprule
\multirow{2}{*}{\textbf{Profile}} & \multicolumn{2}{c}{\textbf{CubeSat}} & \multicolumn{2}{c}{\textbf{Microsat}} & \multicolumn{2}{c}{\textbf{SmallSat}} \\
 & LS & EKF & LS & EKF & LS & EKF \\
\midrule
one step    & \textbf{6.95} & 10.98 & 2.70 & \textbf{1.04} & \textbf{0.86} & 1.73 \\
multi step  & 3.66 & \textbf{1.43} & 1.35 & \textbf{0.54} & 0.97 & \textbf{0.15} \\
sawtooth    & \textbf{4.07} & 6.34 & 2.24 & \textbf{1.56} & 1.30 & \textbf{0.63} \\
sine        & \textbf{3.73} & 5.71 & 0.98 & \textbf{0.44} & 0.30 & \textbf{0.16} \\
multi sine  & \textbf{2.61} & 9.26 & 1.63 & \textbf{1.37} & \textbf{0.68} & 4.72 \\
prbs        & 4.44 & \textbf{2.06} & \textbf{2.08} & 5.16 & \textbf{2.20} & 9.97 \\
chirp       & 2.04 & \textbf{2.04} & 1.70 & \textbf{1.42} & 1.03 & \textbf{0.72} \\
sine-3axis  & \textbf{2.85} & 3.81 & \textbf{0.43} & 3.86 & \textbf{0.23} & 0.31 \\
\bottomrule
\end{tabular}
\end{table}

\begin{table*}[!h]
\centering
\small
\caption{Mean LS and EKF estimation errors (\%) for each torque profile and dynamic inertia mode across three satellite configurations. Best (lowest) per cell is \textbf{bold}.}
\label{tab:dynamic_inertia_errors}
\renewcommand{\arraystretch}{1.2}
\setlength{\tabcolsep}{3pt}
\begin{tabular}{l|cc|cc|cc|cc|cc|cc|cc|cc|cc|}
\toprule
& \multicolumn{6}{c|}{\textbf{CubeSat}} & \multicolumn{6}{c|}{\textbf{Microsat}} & \multicolumn{6}{c}{\textbf{SmallSat}}\\
& \multicolumn{2}{c|}{Step} & \multicolumn{2}{c|}{Drift} & \multicolumn{2}{c|}{Periodic}
& \multicolumn{2}{c|}{Step} & \multicolumn{2}{c|}{Drift} & \multicolumn{2}{c|}{Periodic}
& \multicolumn{2}{c|}{Step} & \multicolumn{2}{c|}{Drift} & \multicolumn{2}{c}{Periodic}\\
& LS & EKF & LS & EKF & LS & EKF & LS & EKF & LS & EKF & LS & EKF & LS & EKF & LS & EKF & LS & EKF \\
\midrule
 One Step & \textbf{3.29} & 11.47 & \textbf{2.61} & 6.85 & 7.27 & \textbf{3.03} & \textbf{6.67} & 9.46 & \textbf{1.80} & 4.71 & 3.15 & \textbf{1.98} & \textbf{8.32} & 10.37 & \textbf{3.50} & 5.67 & \textbf{2.08} & 2.14 \\
 Multi Step & \textbf{5.87} & 6.18 & \textbf{1.23} & 2.53 & \textbf{4.02} & 4.63 & \textbf{7.88} & 8.59 & \textbf{3.04} & 3.79 & 2.28 & \textbf{2.00} & \textbf{8.22} & 9.07 & \textbf{3.40} & 4.29 & 2.12 & \textbf{1.90} \\
 Sawtooth & 5.52 & \textbf{4.88} & \textbf{1.08} & 2.23 & \textbf{4.40} & 5.70 & \textbf{7.08} & 8.01 & \textbf{2.21} & 3.18 & 2.76 & \textbf{2.15} & \textbf{7.93} & 8.66 & \textbf{3.09} & 3.86 & 2.25 & \textbf{1.90} \\
 Sine & 5.81 & \textbf{4.56} & \textbf{1.32} & 2.02 & \textbf{4.09} & 5.84 & \textbf{8.21} & 8.69 & \textbf{3.39} & 3.89 & 2.12 & \textbf{1.94} & \textbf{8.82} & 9.18 & \textbf{4.03} & 4.41 & 1.91 & \textbf{1.84} \\
 Multi Sine & \textbf{6.80} & 8.78 & \textbf{2.21} & 4.14 & 3.16 & \textbf{2.50} & 7.63 & \textbf{7.48} & 2.78 & \textbf{2.78} & \textbf{2.41} & 2.66 & 8.48 & \textbf{7.60} & 3.67 & \textbf{2.77} & \textbf{2.02} & 2.60 \\
 Prbs & 5.18 & \textbf{4.21} & \textbf{0.83} & 8.60 & \textbf{4.76} & 13.57 & \textbf{7.23} & 9.68 & \textbf{2.36} & 4.95 & 2.65 & \textbf{1.97} & 7.12 & \textbf{5.58} & 2.26 & \textbf{0.85} & \textbf{2.74} & 4.27 \\
 Chirp & 7.27 & \textbf{7.12} & \textbf{2.40} & 2.52 & \textbf{2.64} & 3.20 & \textbf{7.56} & 8.26 & \textbf{2.71} & 3.44 & 2.44 & \textbf{2.13} & \textbf{8.17} & 8.81 & \textbf{3.34} & 4.02 & 2.14 & \textbf{1.95} \\
 Sine 3-Axis & 6.56 & \textbf{5.12} & \textbf{1.72} & 2.44 & \textbf{3.27} & 5.91 & 8.71 & \textbf{7.08} & 3.91 & \textbf{2.28} & \textbf{1.94} & 2.89 & \textbf{8.88} & 8.97 & \textbf{4.09} & 4.19 & 1.90 & \textbf{1.88} \\
\bottomrule
\end{tabular}
\end{table*}

\subsection{Experimental Setup}

Experiments are conducted on three satellite configurations (CubeSat, Microsat, SmallSat) under eight excitation profiles (Table~\ref{tab:torque_profiles}). For each configuration, both constant and dynamic inertia cases are tested. Each run spans $T = 300\,\mathrm{s}$ and all results are averaged over 10 random seeds.
To determine a suitable experiment horizon, we compares the average relative error of LS and EKF over durations from 10\,s to 600\,s, aggregated over profiles and seeds. Accuracy improves with longer horizons, with most configurations showing minimal gains beyond 300\,s, which we use as common horizon for all the experiments.

\subsection{Static Inertia Estimation Accuracy}

Table~\ref{tab:static_inertia_grouped} reports static-inertia estimation errors for all torque profiles and satellite configurations. Results reveal distinct trends linked to both excitation structure and estimator assumptions. The batch LS method achieves its best performance with smooth and spectrally rich profiles such as chirp and multi-sine, where persistently exciting yet low-frequency-dominated inputs improve the conditioning of the regression and reduce amplification of measurement noise in the finite-difference computation of angular accelerations. This behavior is beneficial for the smallest satellite, where sensor noise is higher and EKF's reliance on derivative updates makes it more susceptible to measurement variability.

Conversely, EKF performs better under profiles that introduce more abrupt dynamics (multi-step, sine-3axis), especially for larger satellites. In these cases, the satellite’s higher inertia and stronger dynamic coupling yield larger, faster state variations that continuously excite the EKF’s prediction–correction loop, allowing the filter to refine inertia estimates through frequent covariance updates. The interplay between excitation bandwidth, sat inertia, and noise levels determines which estimator is more robust: LS benefits from well-conditioned fits in low-noise regimes, while EKF adapts better to transient information in dynamically rich cases.

Figure~\ref{fig:static_barplots} visualizes these patterns, with LS generally outperforming EKF for smooth inputs and EKF showing advantages for excitations with richer time-domain dynamics. Error bars confirm greater variability for the smallest satellite, consistent with its higher noise sensitivity.

\subsection{Estimation under Time-Varying Inertia}

\begin{figure*}[!h]
     \includegraphics[width=1.\textwidth, height=0.45\textheight]{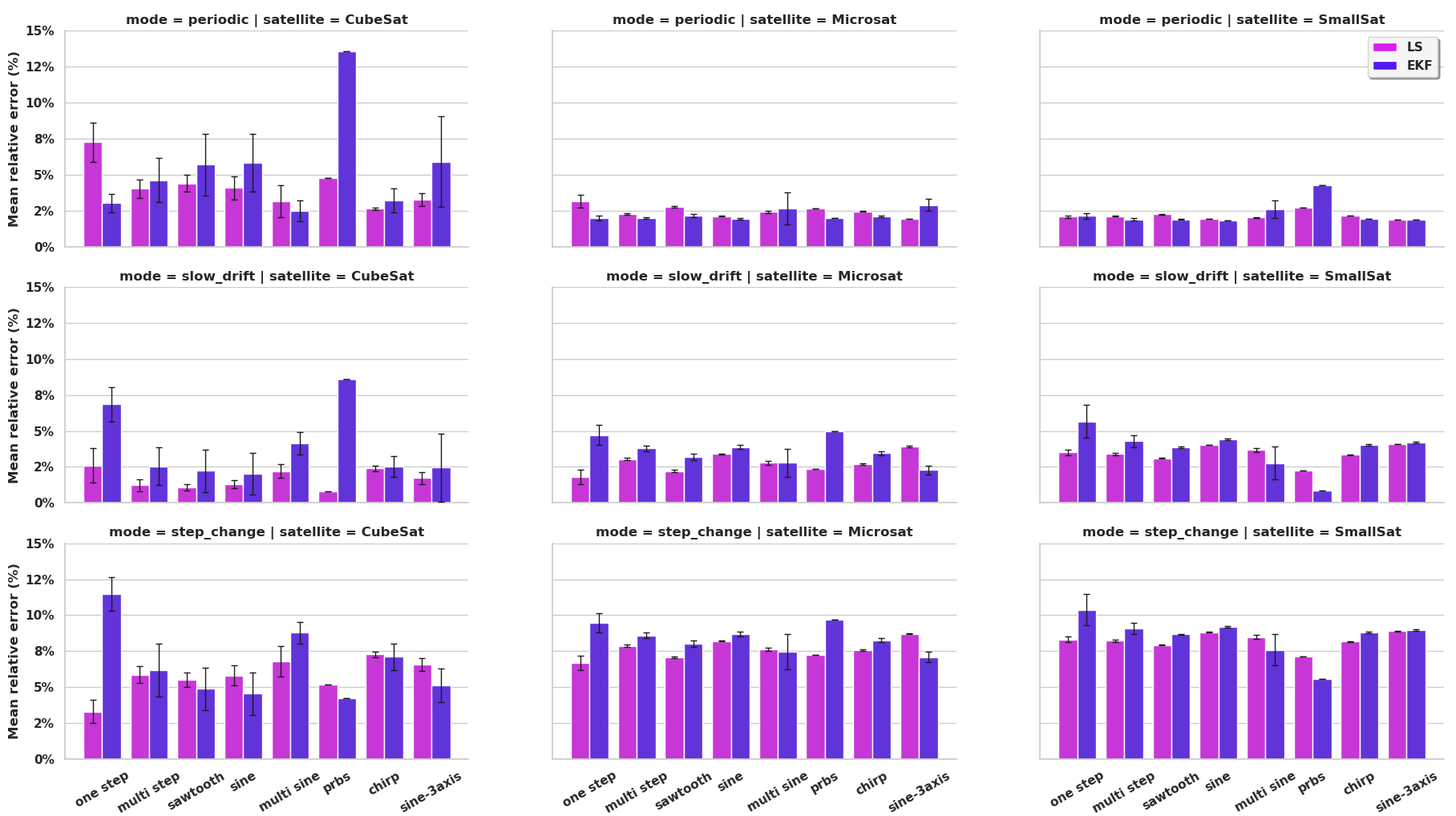}
    \small
\caption{Mean and standard deviation of inertia estimation errors for dynamic scenarios across satellite configurations (columns) and dynamic inertia modality (rows). Results are averaged over 10 seeds, with error bars showing variability.}
    \label{fig:ekf_tracking_dynamic}
    \vspace{-0.6cm}
\end{figure*}

We evaluate performance under three dynamic-inertia modes (\textit{step change}, \textit{slow drift}, \textit{periodic variation}) for all satellites and excitation profiles, with results in Figure~\ref{fig:ekf_tracking_dynamic} and Table~\ref{tab:dynamic_inertia_errors}. EKF generally achieves lower errors than LS for profiles providing persistent, broadband excitation (\textit{multi-step}, \textit{chirp}, \textit{PRBS}), reflecting its ability to track gradual parameter evolution, while LS remains competitive for short or impulsive excitations (\textit{one-step}) where the constant-inertia assumption is adequate. For \textit{multi-sine}, LS performs better on CubeSat but EKF gains on larger satellites, likely due to improved tracking of smoother variations. \textit{Chirp} yields consistently low errors for both methods, while \textit{PRBS} favors EKF except for SmallSat, where LS remains comparable. Across modes, \textit{slow drift} and \textit{periodic} produce similar errors, suggesting excitation richness matters more than variation pattern, whereas \textit{step change} often reduces EKF’s advantage due to transient lag after abrupt changes. Smaller satellites show higher EKF variance, especially for uneven spectra (\textit{multi-sine}, \textit{PRBS}), while larger satellites display more stable trends. Compared to static conditions, LS remains stable for well-conditioned profiles but degrades under gradual changes, EKF matches LS when inertia is constant but surpasses it for smooth variations, and both degrade similarly for abrupt changes. Overall, excitation–estimator pairing should be chosen according to expected inertia variability, satellite size, and operational constraints.

\section{Conclusion}
We presented a comparative analysis of excitation design and estimator behavior for spacecraft inertia identification, focusing on the interaction between torque profile structure, satellite configuration, and estimation method. Our results highlight that smooth and spectrally rich profiles (e.g., chirp, multi-sine) improve regression conditioning and benefit batch Least Squares estimators, particularly in low-noise, small-inertia regimes. Conversely, profiles with abrupt transitions or richer temporal dynamics (e.g., multi-step, 3-axis sine) better exploit the recursive nature of Extended Kalman Filters, especially in higher-inertia configurations where dynamic coupling is stronger.

This study provides actionable guidelines for designing in-orbit excitation strategies and selecting estimation methods based on satellite properties and mission constraints. Our current approach, while still retaining physical validity, works even without enforcing physical constraints on the inertia tensor, making it a valid constraint agnostic solution to dynamic inertia estimation. Future work may incorporate constrained estimation techniques, longer-horizon scenarios, or reinforcement learning-based excitation policies to further enhance robustness and autonomy in on-board inertia identification.


\bibliographystyle{IEEEtran}
\bibliography{bibliography}

\end{document}